\pdfoutput=1

\documentclass[11pt]{article}

\usepackage[final]{acl}

\usepackage{times}
\usepackage{latexsym}

\usepackage[T1]{fontenc}

\usepackage[utf8]{inputenc}

\usepackage{microtype}

\usepackage{inconsolata}

\usepackage{graphicx}

\usepackage{float}     
\usepackage{enumitem}
\usepackage{color}
\usepackage{breqn}
\usepackage{multirow}
\usepackage{amsmath,amsfonts}
\usepackage{adjustbox}
\usepackage{multirow}

\usepackage{hyperref}

\title{Learning to Substitute Words with Model-based Score Ranking}

\author{Hongye Liu \\
  Duke University \\
  \texttt{hongye.liu@duke.edu} \\\And
  Ricardo Henao \\
  Duke University \\
  \texttt{ricardo.henao@duke.edu} \\}

\begin{document}

\maketitle

\begin{abstract}
Smart word substitution aims to enhance sentence quality by improving word choices; however current benchmarks rely on human-labeled data.
Since word choices are inherently subjective, ground-truth word substitutions generated by a small group of annotators are often incomplete and likely not generalizable.
To circumvent this issue, we instead employ a model-based score (BARTScore) to quantify sentence quality, thus forgoing the need for human annotations. 
Specifically, we use this score to define a distribution for each word substitution, allowing one to test whether a substitution is statistically superior relative to others.
In addition, we propose a loss function that directly optimizes the alignment between model predictions and sentence scores, while also enhancing the overall quality score of a substitution.
Crucially, model learning no longer requires human labels, thus avoiding the cost of annotation while maintaining the quality of the text modified with substitutions.
Experimental results show that the proposed approach outperforms both masked language models (BERT, BART) and large language models (GPT-4, LLaMA). The source code is available at \href{https://github.com/Hyfred/Substitute-Words-with-Ranking}{https://github.com/Hyfred/Substitute-Words-with-Ranking}. 
\end{abstract}

\section{Introduction}
In the current era of AI-driven technologies, the generation of natural language by machines has become a critical area of research within the field of natural language processing (NLP).
The advent of advanced language models such as GPT \citep{ouyang2022training} and LLaMA \citep{touvron2023llama, touvron2023llama2, dubey2024llama} has brought about unprecedented improvements in generating coherent, contextually accurate text across a wide range of applications, including machine translation, summarization, and conversational agents.
These models are revolutionizing industries by enabling machines to produce human-like text at scale, greatly expanding the possibilities for automation and interaction in both professional and creative domains.

As language generation capabilities improve, the need to refine and control the output of these models becomes more pressing.
One key area in this direction is smart word suggestion (SWS) defined by \citet{wang2023smart} as a task that focuses on enhancing writing by improving two aspects of NLP: identifying words (or tokens more generally) that if replaced can improve sentence quality, and suggesting alternatives for such words. 
Complementarily, \citet{wang2023smart} introduced a dataset to benchmark models built for SWS.
This task is essential in applications such as paraphrasing, machine translation, and writing assistance, where accurate word choice directly impacts the clarity and quality of the generated text.
Despite the progress in full-text generation, word substitution presents unique challenges in maintaining grammatical correctness, contextual appropriateness, and semantic fidelity.

Previous SWS work can be divided into {\em task-specific} models and {\em prompt-based} methods.
Task-specific models are trained directly to address the SWS task.
One approach adapts models for lexical substitution (LS) \citep{mccarthy2007semeval, kremer2014substitutes}, allowing them to generate substitutions for each word in a sentence.
If the substitution differs from the original word, it is recognized and recorded as such.
BERT-based \citep{zhou2019BERT} and the LexSubCon \citep{michalopoulos2021lexsubcon} models were first explored in \citet{wang2023smart}.
They also fine-tuned pretrained models like BERT \citep{devlin2018BERT} and BART \citep{lewis2019bart}, allowing them to handle SWS end-to-end.
Importantly, these models were fine-tuned using supervised learning to provide suggestions for Wikipedia text using ground-truth substitutions obtained from a list of synonyms gathered from a thesaurus.
Orthogonal to these machine learning techniques, rule-based methods that rely on paraphrasing databases ({\em e.g.}, PPDB \citep{pavlick2015ppdb}) or a thesaurus ({\em e.g.}, the Merriam-Webster thesaurus) have also been explored \citep{wang2023smart}.

Alternatively, prompt-based techniques leveraging large language models (LLMs), such as GPT \citep{ouyang2022training}, have gained popularity in SWS.
Provided that research in language generation has shifted from narrow tasks like substitution to generating coherent, contextually rich text, tasks such as SWS have become less of a priority.
However, in general it is still key to enhance sentence quality, as precise word choices can significantly impact clarity, tone, and overall effectiveness in communication.
Moreover, the success of these methods is highly dependent on prompt design.

From a different perspective, benchmarks for SWS predominantly rely on human-labeled data for both training and evaluating model performance.
However, only a small group of annotators are recruited for data labeling, which can lead to a lack of diversity.
For instance, in the SWS test dataset \citep{wang2023smart}, 12,883 suggestions for target words are available of which 7,827 (60.8\%) were created by a single annotator, only 1632 (12.67\%) of suggestions were created by at least three annotators, and merely 633 (4.91\%) of suggestions were created by more than three annotators.
While scaling up the number of annotators could potentially mitigate this issue, it significantly increases the costs of generating the annotations.
In the context of substitution tasks, a large number of annotators is necessary to obtain a comprehensive representation of the substitution space, as evaluating them requires a diverse sample of high-quality alternatives.
As a result, validation sets in such benchmarks may not fully capture model performance.
Conversely, recent research has shown that model-based evaluation metrics align closely with human judgments \citep{lu2022toward, yuan2021bartscore, he2024improving}, offering a viable alternative to alleviate the drawbacks of human evaluations.

To address these challenges, we propose an approach that eliminates the dependency on human-labeled data.
Our contributions are listed below:
\begin{itemize}[leftmargin=10pt,itemsep=0pt,topsep=0pt,parsep=0pt]
    \item We identified a gap in the evaluation methods for substitution models, which rely on human annotators.
    We propose a model-based evaluation strategy based on the BARTScore to assess SWS models without human annotations.
    \item We introduce an approach for generating and ranking token substitutions.
    Our method optimizes both the quality of the substitutions, via the model-based score, and the ranking of the substitutions relative to such a score.
    To the best of our knowledge, this is the first time SWS has been tackled by jointly optimizing the association between predictions and quality scores.
    \item We redefined the substitution task by forgoing the need for supervised learning relying on human annotations.
    Thus, our method eliminates the need for costly human-labeled data while maintaining high performance in SWS tasks.
    \item Our experimental results demonstrate that the proposed approach outperforms both masked language models (BERT, BART) and LLMs (GPT-4, LLaMA) in the SWS task.
\end{itemize}

\section{Related Work}
Enhancing word usage is a key feature of writing assistance.
The SWS task in \citet{wang2023smart} addresses it by emulating a real-world end-to-end writing scenario.
Unlike traditional LS tasks \citep{mccarthy2007semeval, kremer2014substitutes}, SWS requires systems to identify words that, if replaced, can improve sentence quality and suggest alternatives for such words.
In the LS task, however, the word to be replaced is predefined and the model is only tasked with finding suitable substitutions without altering the overall meaning of the sentence.
Further, in LS substitutions are typically lemmatized to match ground-truth labels, but in SWS they must be in the correct grammatical form.

\citet{zhou2019BERT} proposed a LS method using contextual word embeddings, modifying BERT with a dropout embedding policy to partially mask target words and introduced a validation score based on the top four layers of BERT for candidate evaluation.
\citet{michalopoulos2021lexsubcon} integrated external knowledge from WordNet into BERT for LS, combining scores from BERT, WordNet gloss similarity, and sentence embeddings, along with the validation score from \citet{zhou2019BERT}, to generate substitutes.
ParaLS \citep{qiang2023parals} generated substitutes through a paraphrasing model, using a heuristics-based decoding strategy.

To adapt the substitution capabilities from LS to the SWS task, \citet{wang2023smart} used an approach that allowed models to generate substitutions for each word.
If the substitution differs from the original word, it is recognized as the word the model wants to improve and recorded as a suggestion.
Further, they fine-tuned pretrained models like BERT \citep{devlin2018BERT} and BART \citep{lewis2019bart} to address the task in an end-to-end manner, and as an alternative to machine learning, rule-based methods using dictionaries (thesaurus) were also explored.

Recently, prompt-based methods have emerged as a powerful alternative for SWS \citep{wang2023smart}, by leveraging the capabilities of LLMs such as GPT \citep{ouyang2022training}.
These approaches shift the focus from traditional model fine-tuning to designing prompts that guide the model to produce desired outcomes.
While this flexibility allows for easier adaptation to various tasks, the effectiveness of the method heavily relies on the prompt's structure.
Despite the growing interest in generating fluent, contextually cohesive text with LLMs, prompt-based methods often overlook the importance of improving specific word choices.
In contexts where sentence clarity and precision are key, refining word choices remains an important task that prompt-based approaches need to address.

Compared to traditional methods that rely heavily on human involvement, {\em e.g.}, via manual annotations, in our work, we utilize model-based sentence scoring, specifically, BARTScore to quantify the quality of sentences, thereby eliminating the human-associated costs.
We align the model’s output with the scores through a ranking loss, ensuring that substitutions with higher scores are more likely to be generated by the model, resulting in higher-quality substitutions.
This approach forgoes the need for human-dependent supervised training, thus without the need for human intervention.

\section{Methodology}
In SWS, achieving a representative sample of high-quality alternatives requires a broad pool of annotators, as diverse substitution options are essential for accurate evaluation.
However, relying solely on human annotators is thus not only costly but also inefficient. 
Below we introduce a method to train and evaluate token substitution models that does not require human annotations, by instead solely relying on model-based scoring functions.

\paragraph{Problem Definition}
Let $X$ denote a sentence composed of $N$ tokens, {\em i.e.}, $X=(x_1,\ldots,x_N)$.
We aim to find a collection of $K$ potential substitutes $\{\tilde{w}_k\}_{k=1}^K$ for token $x_n$, while estimating their likelihood given the sentence ({\em i.e.}, its context) as $p_\theta(x_n=\tilde{w}_k|X)$, where $X$ is the original sentence and $p_\theta(\cdot)$ is a model for the conditional likelihood for token $x_n$ parameterized by $\theta$.
Note that technically, if $p_\theta(\cdot)$ is a masked language model we should write $p_\theta(x_n=\tilde{w}_k|X_{\backslash x_n})$, where $X_{\backslash x_n}$ indicates that $x_n$ has been masked out in $X$, and alternatively, if we use an (autoregressive) language model we should write $p_\theta(x_n=\tilde{w}_k|x_{<n},Z)$, where $x_{<n}$ is the portion of the original sentence up to token $n-1$ and $Z$ is a prompt.
However, we use $p_\theta(x_n=\tilde{w}_k|X)$ in the following for notational simplicity.
Based on these likelihood estimates, the token substitution rule is set as $x_n \leftarrow \tilde{w}_k$ if $p_\theta(x_n=\tilde{w}_k|X)>p_\theta(x_n=w_n|X)$, where $w_n$ is the original value of $x_n$.
Moreover, if multiple replacement candidates satisfy the substitution rule, we select $\tilde{w}_k$ according to ${\rm argmax}_k \ p_\theta(x_n=\tilde{w}_k|X)$, and alternatively, if none satisfy it, we leave the token unchanged, {\em i.e.}, $x_n \leftarrow w_n$.

Importantly, $p_\theta(x_n=\tilde{w}_k|X)$ reflects the likelihood of token $\tilde{w}_k$ given its context $X$, rather than the quality of the sentence.
So motivated, we also seek to align such estimates with a score $M(x_n=\tilde{w}_k|X)$ quantifying the quality of sentence $X$ with $x_n$ substituted with $\tilde{w}_k$.
For simplicity, in the following we denote the sentence $X$ with $x_n$ replaced with the $k$-th candidate, $\tilde{w}_k$, as $\tilde{X}_{k}=(\tilde{x}_1,\ldots,\tilde{x}_N)$.
Further, to make our objective scalable and general, the score $M(x_n=\tilde{w}_k|X)$ ought not be obtained via human feedback, but via a {\em black-box model} with which we can score sentences, but through which we cannot learn, {\em i.e.}, gradients cannot be propagated through it to obtain learning signals.

Below we start by showing how we can use a model-based score to statistically characterize the suitability of a candidate as an alternative to the substitution rule introduced above, and then present an approach to estimate $P(x_n=\tilde{w}_k|X)$ while accounting for scores $M(x_n=\tilde{w}_k|X)$.

\subsection{Statistic for Model-based Scores}\label{sc:model-based_stat}
\paragraph{Model-based Score}
Using automated scoring methods to quantify text generation quality without relying on human annotators can reduce costs associated with human labeling. 
Model-based scoring approaches such as BLEURT \citep{sellam2020bleurt}, BERTScore \citep{zhang2019BERTscore}, GPTScore \citep{fu2023gptscore}, and BARTScore \citep{yuan2021bartscore}, have proven to be effective while closely aligning with human evaluation results.
BARTScore is one of the most popular metrics for the evaluation of text generation.
It has been shown that BARTScore outperforms other metrics such as BERTScore and BLEURT \citep{yuan2021bartscore}, while also being more efficient than GPTScore.
Further, it has been used also as a ranking tool for substitution tasks, achieving good results \citep{qiang2023parals}.

The BARTScore is a metric for universal natural language generation evaluation \citep{yuan2021bartscore}.
It leverages the conditional likelihood from a pre-trained BART model \citep{lewis2019bart} to assess the quality of generated text.
Since BART is an autoregressive model, the log-likelihood of tokens in a sentence is obtained one at a time conditioned on tokens before it.
Specifically, we write
\begin{equation}\label{eq:bartscore}
M(\tilde{X}_k)=\textstyle{\sum}_{n=1}^{N} \log p_{\hat{\theta}}\left(\tilde{x}_{n} \mid \tilde{x}_{<n}, X\right) \,,
\end{equation}
where $\hat{\theta}$ represents the parameters of a pre-trained model (in this case, BART\footnote{We use \href{https://github.com/neulab/BARTScore}{BARTScore (ParaBank2 version)} without incorporating any prompts.}), $\tilde{x}_n$ is the $n$-th token of the modified sentence $\tilde{X}_k$ which we wish to score, $\tilde{x}_{<n}$ represents the first $n-1$ tokens of $\tilde{X}_k$, and $X$ is the original (unmodified) sentence.
In \eqref{eq:bartscore}, the BARTScore $M(x_n=\tilde{w}_k|X)$ has been simplified for notational convenience to $M(\tilde{X}_k)$, {\em i.e.}, the score of modified sentence $\tilde{X}_k$ in which a token has been replaced with the $k$-th candidate from potential substitutes $\{\tilde{w}_k\}_{k=1}^K$.
For now, we assume that such candidates are readily available, however, in the next section we will propose a learning strategy to train a model to generate substitutes given $X$.

\paragraph{Model-based Score Statistic}
One potential issue with model-based scores including the BARTScore in \eqref{eq:bartscore} is that there is no clear standard to determine what score (or score range) qualifies as ``good''.
It is clear that though scores can be used for pairwise or aggregate comparisons, the magnitude of the difference between two scores is not meaningful without a reference score distribution.

This underscores that numerical scores alone are insufficient to fully capture the significance of a score difference.
Inspired by statistical hypothesis testing \citep{moore1999bootstrapping}, we propose constructing a reference (null) distribution of $M(\tilde{X}_k)$ by sampling a collection of substitutes for a given token and then calculating their model-based score.
By determining where \(M(\tilde{X}_k)\) lies within this empirical distribution, we can more reliably assess the quality of a substitute.
Since sampling candidates uniformly at random from a natural language vocabulary or from empirical token frequencies will be hugely inefficient, instead we use a model to generate substitute candidates.
In this manner, we can evaluate the relative quality of the generated substitutes compared to other candidates from the model’s own distribution.
This constitutes a principled framework for assessing the quality of a model in generating token substitutes.
Next, we introduce a hypothesis test to estimate a $p$-value for the significance of a candidate substitute relative to a reference (empirical) substitute distribution.

Given a model we wish to evaluate, let $\{\tilde{w}_k\}_{k=1}^{K_s}$ be a set of $K_s$ candidates for sentence $X$ at a given position produced by such model, {\em e.g.}, a masked language model like BERT \citep{devlin2018BERT}.
Note that in general, $K_s$ is a number much larger than the size of potential substitutes $K$ described before.
We can evaluate the quality of, for instance, the top candidate $w_1$ (assuming they are ordered) as the proportion of times it produces a score that is larger than using all other candidates, $\{\tilde{w}_k\}_{k=2}^{K_s}$.
This frequency or $p$-value is formally written as
\begin{equation}
p_{\tilde{X}_1} = \frac{1}{K_s-1} \textstyle{\sum}_{k=2}^{K_s} \ I[M(\tilde{X}_k)>M(\tilde{X}_1)] \,,
\label{eq:p-value}
\end{equation}
where $I[\cdot]$ is the indicator function and the $\{w_k\}_{k=1}^{K_s}$ producing modified sentences $\{\tilde{X}_k\}_{k=1}^{K_s}$ are obtained from a model $p_\theta(x_n=\tilde{w}_k|X)$, which defines the reference distribution.
In standard hypothesis testing fashion, we reject the hypothesis that candidate $w_1$ has the same expected score as the reference distribution if $p_{\tilde{X}_1}<\alpha$, for a significance level $\alpha$, which we set to $\alpha=0.01$ in our experiments.
Since calculating model-based scores such as BARTScore is computationally expensive, we trade-off statistical accuracy with computational efficiency and let $K_s=1000$ in the experiments.

\subsection{Preference-Aware Learning}\label{sec:Preference-Aware-Learning}
Now that we have constructed a statistic based on model-based scores, in principle, we seek to train a model such that for each token $x_{n}$, the resulting $p_{\tilde{X}_k}$ from \eqref{eq:p-value} for a given substitute candidate $w_k$ producing $\tilde{X}_k$ is as small as possible.
According to \eqref{eq:p-value}, it is desirable for $M(\tilde{X}_{1})$ to be larger than $M(\tilde{X}_{\hat{k}})$ for $\{w_k\}_{k=1}^K$ candidates sampled from model $p_\theta(x_n|X)$.
This means that effectively we require the model being such that its outputs align, in likelihood, with the model-based score, BARTScore here.
Consequently, it is desirable to learn parameters $\theta$ so $p_\theta(x_n=w_k|X)$ is ranked consistent to $M(\tilde{X}_{k})$, which in turn will make $p_{\tilde{X}_k}\to 0$ in \eqref{eq:p-value} for $k\to 1$.
Below we consider two approaches to accomplish this: margin ranking \citep{liu2023learning, chern2023improving, liu2022brio} and direct preference optimization (DPO) loss \citep{rafailov2024direct}.
%

\paragraph{Ranking Optimization}
Ranking has been used as an optimization objective in many tasks including text summarization \citep{chern2023improving, liu2022brio}.
Maximum likelihood estimation based on the standard cross-entropy loss can be effective at satisfying \eqref{eq:p-value} by setting it as a classification problem by defining a vector of labels $y$ where $y_k=1$ if $M(\tilde{X}_{k})$ is maximum among $\{M(\tilde{X}_{k})\}_{k=1}^K$ or $y_k=0$ otherwise.
However, it does not take into account the ordering of the model-based scores or their magnitude differences.
So motivated, we consider the following margin ranking (MR) loss \citep{liu2023learning, chern2023improving, liu2022brio}
\begin{equation}
\mathcal{L}_{\rm MR} = \textstyle{\sum}_{k}^{K} \textstyle{\sum}_{j > k}^{K} \ \max(0, s_j - s_k + \lambda_{jk}) \,,
\label{eq:constra}
\end{equation}
where $\{s_k\}_{k=1}^K$ are the logits from model $p_\theta(x_n=w_k|X)$, {\em i.e.}, before applying the softmax function, and we have sorted them such that $s_k>s_j$ if $M(\tilde{X}_{k}) > M(\tilde{X}_{j})$ for $i,j=1,\ldots,K$.
Further, we set the {\em margin} $\lambda_{jk}=\lambda\times(j-k)$ for some hyperparameter $\lambda$, which in the experiments is set via cross-validation.
Intuitively, \eqref{eq:constra} encourages the model to make predictions whose outputs $\{p_\theta(x_n=w_k|X_{\backslash x_n})\}_{k=1}^K$ are consistent in order with $\{M(\tilde{X}_{k})\}_{k=1}^K$, by penalizing pairwise order mismatches, while also enforcing predictions to be distanced by a fixed margin to improve robustness.
The latter is justified by extensive results from the margin learning literature \citep{smola2000advances}.

\paragraph{Improving Model-based Scores}
One unintended consequence of the loss function in \eqref{eq:constra} is that though it encourages model predictions to be aligned with model-based scores, it does so regardless of their values.
This is so because only the order of $\{M(\tilde{X}_{k})\}_{k=1}^K$ is considered in \eqref{eq:constra}.
In practice, we have observed that \eqref{eq:constra} effectively improves the ranking of predictions from the model, but it does so at the expense of producing predictions that have, on average, lower model-based scores relative to a reference model, {\em e.g.}, the pre-trained model used as initialization for the refinement of $\{p_\theta(x_n=w_k|X)\}_{k=1}^K$.
To address this issue, we consider two approaches, one that seeks to improve the weighted average of model-based scores and another that maximizes the model-based score of the top prediction from the model relative to that of the reference model.
Specifically, we write
\begin{align}
    \mathcal{L}_{\rm AS} & =-\textstyle{\sum}_{k}^{K} \ h(s_{k}) M(\tilde{X}_k) \,, \label{eq:as} \\
    \mathcal{L}_{\rm BS} & =\operatorname{max}(0,(M(X)-M(\tilde{X_{1}}))f\left(s_{1}\right)) \,, \label{eq:bs} 
\end{align}
where $h(\cdot)$ is the softmax function, $s_k$ is the logit corresponding to $p_\theta(x_n=w_k|X)$, and $M(X)$ and $M(\tilde{X_{k}})$ are the model-based scores of the original and modified sentence, respectively.
Recall that for $M(\tilde{X_{1}})$, the token of interest $x_n$ in $X$ has been modified to the top prediction $w_1$ from the model.
Conceptually, $\mathcal{L}_{\rm AS}$ in \eqref{eq:as} seeks to maximize the (weighted) average of model-based scores from $K$ predictions, while $\mathcal{L}_{\rm BS}$ aims to improve the model-based score of the top prediction with respect to that of a reference model.

We then combine the margin ranking loss in \eqref{eq:constra} with the score-improving losses in \eqref{eq:as} or \eqref{eq:bs} as
\begin{align}
    \mathcal{L}_{\rm MR+AS} & =\mathcal{L}_{\rm MR} + \gamma\mathcal{L}_{\rm AS} \,, \label{eq:mr_as}\\
    \mathcal{L}_{\rm MR+BS} & =\mathcal{L}_{\rm MR} + \gamma\mathcal{L}_{\rm BS} \,, \label{eq:mr_bs}
\end{align}
where $\gamma$ is a hyperparameter trading off ranking or model-based score improvement.
In the experiments we will compare these two approaches to determine empirically whether is better to attempt to improve the scores of $K$ predictions from the model as opposed to improving the score of one of the predictions from the model (the most likely) relative to a baseline prediction obtained from a reference model.
Note that is tempting to use \eqref{eq:bs} with all predictions, not just the first one, however, we found empirically that it considerably increases the computational cost without significant performance gains.
The cost overhead is caused mainly by the need to evaluate the model-based score $K$ times rather than just $2$ in \eqref{eq:bs}. 

\paragraph{Direct Preference Optimization}
Direct Preference Optimization (DPO) \citep{rafailov2024direct} is an efficient technique for aligning large language models with human feedback, which gained popularity due to its simplicity \citep{miao2024inform}.
For instance, it has demonstrated to be effective in chat benchmarks \citep{tunstall2023zephyr, zheng2023judging}.
In our case, the model-based score serves as proxy for human feedback
DPO (under the Plackett-Luce model) and is formally expressed as
\begin{align}\label{eq:dpo}
\mathcal{L}_{\rm DPO} = -\mathbb{E} 
\left[ \log \textstyle{\prod}_{k=1}^{K} 
\frac{\exp ( \delta  r_k )}{ \sum_{j=k}^{K} \exp ( \delta r_j )} \right] \,, 
%
\end{align}
where $r_k=\log p_\theta(\tilde{X}_k) - \log p_{\hat{\theta}}(\tilde{X}_k)$, the expectation is over $\{\tilde{X}_1, \ldots, \tilde{X}_K, X\}$, and 
$\theta$ and $\hat{\theta}$ denote the parameters of the model being trained and that used for reference, respectively.
Accordingly, only $\theta$ are updated while learning while $\hat{\theta}$ are kept fixed.

Since the magnitude of the logits drops significantly as $k\to K$, we found that the sum in the denominator of \eqref{eq:dpo} weakens the loss.
Therefore, we removed the sum and instead compared the $k$-th and the $(k+1)$-th substitution, rather than comparing it with all $K$ values.
Further, we approximate $\log p_\theta(\tilde{X}_k)$ with its logit $s_k$, and let $\delta=1$ for convenience.
Then \eqref{eq:dpo} simplifies (for a single token in sentence $X$) to 
\begin{align}
    \hspace{-2mm}\mathcal{L}_{\text{DPO*}} &= -\textstyle{\sum}_{k=1}^{K-1} \left( s_k - \hat{s}_k - s_{k+1} + \hat{s}_{k+1} \right) \,, 
    \label{eq:dpo*}
\end{align}
where $s_k$ and $\hat{s}_k$ denote the logit of the $k$-th substitution from the model being trained and the reference model, respectively.

We also extend DPO (under the Bradley-Terry model) \citep{rafailov2024direct} to multiple substitute candidates, where each candidate is compared with the next in the ordered list of substitute candidates.
We write (for a single token in sentence $X$)
\begin{align}
& \mathcal{L}_{\text{$\sigma$DPO*}} = \label{eq:dpo*-sigma} \\
& \hspace{8mm} -\textstyle{\sum}_{k=1}^{K-1} \log\sigma \left( s_k - \hat{s}_k - s_{k+1} + \hat{s}_{k+1} \right) \,, \notag
\end{align}
%
%
%
%
from which we see that the only difference between \eqref{eq:dpo*-sigma} and \eqref{eq:dpo*} is that the comparison of logit values in the former is scaled with the log-logistic function.
See Appendix~\ref{sc:dpo_det} for a derivation of both losses.

\section{Experiments}
%
%
Below we illustrate the problem with evaluating with human annotations, then we present an ablation study comparing the optimization approaches in Section~\ref{sec:Preference-Aware-Learning} and a benchmark comparing the proposed model to MLM and LLM approaches in terms of model-based score alignment, average score and the statistic in \eqref{eq:p-value}. 
Further, we explore the performance of top-2 predictions, {\em i.e.}, when the top prediction is the original token, and results comparing LLMs with and without prompts encouraging token substitutes to be ranked by quality.
%
%

\paragraph{Datasets}
We consider four datasets.
The SWS dataset by \citet{wang2023smart} consists of three sets, validation and test labeled by human annotators, and an artificially generated training set.
The training dataset was constructed from Wikipedia sentences, with labels obtained using a combination of PPDB \citep{pavlick2015ppdb} and the Merriam-Webster thesaurus.
We also consider two traditional lexical substitution datasets, LS07 \citep{mccarthy2007semeval} and LS14 \citep{kremer2014substitutes}, which use lemmatized human annotations as ground truth substitutions.
Further, we also consider a general dataset without ground-truth substitutions.
Specifically, we use XSum \citep{narayan2018don}, a summarization dataset consisting of BBC news articles and corresponding summaries.
In the Appendix we show Table~\ref{tb:data_summ} summarizing these four datasets.

\paragraph{Baselines}
We compare our method with three classes of baseline models.
{\em Masked language models (MLMs):} BERT-base-uncased with original MLM head (BERT-naive) \citep{devlin2018BERT}.
BERT-spsv \citep{zhou2019BERT}, as a representative LS model.
BERT-SWS and BART-SWS, both of which were fine-tuned on the SWS training dataset \citep{wang2023smart}.
{\em Rule-based models:} The rule-based approach introduced by \citet{wang2023smart}, which leverages a thesaurus.
%
%
%
{\em Prompt-based large language models (LLMs):} We leverage the power of pretrained LLMs via prompts to generate token substitutions.
We consider two popular choices, GPT \citep{ouyang2022training} and LLaMA \citep{touvron2023llama, touvron2023llama2, dubey2024llama},
specifically, GPT-4o and LLaMA-3.1-8B-Instruct.
Prompt-based techniques have gained popularity due to their flexibility and effectiveness in generating high-quality, contextually appropriate substitutions without requiring task-specific tuning \citep{liu2023pre}.
We designed both ranking and non-ranking prompts, which we show in Appendix~\ref{sc:llm_prompts}.
%

\paragraph{Evaluation Metrics}
We use the cosine similarity (CS) to measure the correlation between model predictions and model-based scores, {\em i.e.}, BARTScores.
Since BARTScore produces log-likelihoods, we log-transform model predictions accordingly.
We also consider the quality of the substitutes $\{\tilde{w}_k\}_{k=1}^K$.
First we get the score ratio between the sentence with and without substitution and then calculate the average of BARTScore ratios as $\text{ABR}=1/K\sum_{k=1}^{K}M(\tilde{X}_{k})/M(X)$.

\paragraph{Implementation}
%
For our model, we fine-tuned the BERT-base-uncased model \citep{wang2023smart} on 100k randomly sampled sentences from the SWS training dataset using one Nvidia A100 GPU.
For each sentence, we randomly selected 5 tokens, and for each token, we chose $K=5$ candidates from the model’s output to form a candidate pool and compute the training loss.
The model was fine-tuned over 5 epochs.
For DPO, we duplicated the same BERT model and weights at initialization, freezing one as the reference while only updating the weights of the other during training.
During testing, we used the same token selection strategy, randomly selecting 5 tokens per sentence and choosing the top 5 candidates for each.
Hyperparameter settings are described in Appendix~\ref{sc:hyperparams}.

\begin{table}[t]
\centering
\scalebox{0.75}{
\begin{tabular}{c|ccc}
    & Not change & \multicolumn{1}{c}{Agreement} & Disagreement \\ 
    \hline
    Change & 4125 (0.74) & \multicolumn{1}{c}{631 (0.11)} & 830 (0.15) \\
    Not change & 14138 (0.94) & \multicolumn{2}{c}{969 (0.06)}
\end{tabular}
}
\vspace{-2mm}
\caption{Token changes in SWS test data.
Rows are for annotator changes and columns indicate model changes and if they are in agreement with the annotator.
In parenthesis are proportions relative to the row totals.
}
\label{Sws_test_data_stat}
\vspace{-2mm}
\end{table}

\paragraph{Illustrative Example}
%
%
%
We stratify tokens in the SWS test set into five groups according to whether they were substituted by human annotators and/or the model.
These groups are:
change agreement (CA: both model and the human annotator agreed on the change),
change disagreement (CD: both model and the annotator suggested a change, but the model substitution did not match that of the annotator),
no change agreement (NCA: the model and the annotator kept the token unchanged),
only model changed (OMC: the model suggested a change, but the annotator did not find it necessary), and
only annotator changed (OAC: the annotator suggested a change, but the model did not).

Results in Table~\ref{Sws_test_data_stat} show that most tokens remain unchanged (NCA: 94\%) by both model (BERT-SWS) and annotator.
The proportion of annotator changes for which the model agrees or disagrees is similar, CA: 11\% {\em vs}. CD: 15\%, respectively.
Interestingly, most annotator changes are not captured by the model (OAC: 74\%), which is consistent with results in \citet{wang2023smart} for a variety of models, and OMC: 6\% tokens not changed by the annotator are changed by the model.
These results underscore that $i$) a model is unlikely to match human annotations when these are largely subjective and incomplete; and $ii$) cases in which changes made by the model do not match the human annotator cannot be considered errors.
These are consequences of having a few annotators each selecting which tokens need to be changed.


This is further illustrated below in Table~\ref{table_benchmark} where we show that though tokens for which both model and annotator agree have better model-based statistics, there is also a non-insignificant portion for which model and annotator disagree, but the model-based statistic suggests that the model changes are of good quality.
Some of these examples are shown in Table~\ref{tb:cases} of the Appendix.

\begin{table}[t]
\centering
\adjustbox{width=\columnwidth}{
\begin{tabular}{c|cc|cc|cc|cc|cc}
            & \multicolumn{2}{c|}{SWS}      & \multicolumn{2}{c|}{LS07}     & \multicolumn{2}{c|}{LS14}     & \multicolumn{2}{c|}{XSum}     & \multicolumn{2}{c}{AVG}                                                                                       \\
            & CS            & ABR           & CS            & ABR           & CS            & ABR           & CS            & ABR           & CS                                                                 & ABR                                      \\ \hline
MR          & \textbf{0.99} & 0.82          & 0.93          & 0.81          & 0.93          & 0.78          & 0.93          & 0.79          & 0.94$\pm$0.024                                                     & 0.8$\pm$\textbf{0.012}  \\
DPO                               & 0.91          & 0.87          & 0.9           & 0.86          & 0.91          & 0.83          & 0.85          & 0.82          & 0.89$\pm$0.025                                                     & 0.85$\pm$0.025                           \\
DPO*                              & 0.92          & 0.87          & 0.91          & 0.86          & 0.91          & 0.82          & 0.87          & 0.83          & 0.9$\pm$0.019                                                      & 0.85$\pm$0.021                           \\
$\sigma$DPO* & 0.93          & 0.89          & 0.93          & 0.86          & 0.92          & 0.84          & 0.88          & 0.84          & 0.92$\pm$0.021  & 0.86$\pm$0.02                            \\
MR+BS       & 0.94          & \textbf{0.90} & 0.95          & \textbf{0.87} & 0.95          & \textbf{0.86} & 0.95          & \textbf{0.86} & 0.94$\pm$0.005                                                     & \textbf{0.87}$\pm$0.015 \\
MR+AS       & 0.98          & 0.890         & \textbf{0.99} & 0.85          & \textbf{0.99} & 0.83          & \textbf{0.99} & 0.84          & \textbf{0.99}$\pm$\textbf{0.002} & 0.86$\pm$0.021                          
\end{tabular}}
\vspace{-2mm}
\caption{
Ablation CS and ABR metrics.
Figures for each dataset are medians over all token predictions.
The last column shows averages over all datasets with standard deviations.
The best results are highlighted in {\bf bold}.
}
\label{table_training}
\end{table}

\paragraph{Ablation Study}
%
%
Section~\ref{sec:Preference-Aware-Learning} introduced two approaches to align the predictions of a trained model with model-based scores, namely MR in \eqref{eq:constra} and DPO in \eqref{eq:dpo}, \eqref{eq:dpo*} and \eqref{eq:dpo*-sigma}.
For the former, we consider two ways of improving the model-based scores, via a weighted average in \eqref{eq:mr_as} and improvement relative to a reference model in \eqref{eq:mr_bs}.
Results in Table~\ref{table_training} show results comparing $i$) the alignment of model predictions with model-based scores using median CS, and $ii$) the average model-based scores using median ABR.
Distributions of CS and ABR for all tokens are shown in the Appendix Figures~\ref{fig:bench_cs} and \ref{fig:bench_abr}.
We see that
$i$) the MR loss alone is not sufficient to deliver the best CS and produces the worst ABR; 
$ii$) DPO variants improve the ABR relative to MR, but not in terms of CS;
$c$) MR+BS and MR+AS produce the best overall ABR and CS, respectively, however, MR+AS seems to provide the best trade-off between the two performance metrics, while outperforming both MR and DPO.
Consequently, in the following experiments we will focus on MR+AS.

\begin{table}[t]
\centering
\scalebox{0.47}{
\begin{tabular}{c|cc|cc|cc|cc|cc}
                   & \multicolumn{2}{c|}{SWS}      & \multicolumn{2}{c|}{LS07}     & \multicolumn{2}{c|}{LS14}     & \multicolumn{2}{c|}{XSum}     & \multicolumn{2}{c}{AVG}                                                                                        \\
                   & CS            & ABR           & CS            & ABR           & CS            & ABR           & CS            & ABR           & CS                                                                 & ABR                                       \\ \hline
BERT-naive      & 0.92          & 0.87          & 0.92          & 0.83          & 0.92          & 0.82          & 0.93          & 0.81          & 0.92$\pm$0.003                                                     & 0.83$\pm$0.018                            \\
BERT-spsv         & 0.93          & 0.82          & 0.93          & 0.80          & 0.93          & 0.74          & 0.93          & 0.74          & 0.93$\pm$0.002                                                     & \multicolumn{1}{l}{0.78$\pm$0.038}        \\
BERT-SWS          & 0.92          & 0.84          & 0.92          & 0.81          & 0.93          & 0.78          & 0.93          & 0.79          & 0.93$\pm$0.003                                                     & 0.81$\pm$0.02   \\
BART-SWS         & 0.93          & 0.84          & 0.92          & 0.83          & 0.92          & 0.77          & 0.93          & 0.80          & 0.93$\pm$0.003                                                     & \multicolumn{1}{l}{0.81$\pm$0.028}        \\
Rule-based        & 0.93          & \textbf{0.90}          & 0.92          & 0.83          & 0.93          & 0.82          & 0.93          & 0.83          & 0.93$\pm$0.001                                                     & \multicolumn{1}{l}{0.85$\pm$0.024}        \\
GPT-4o             & 0.95          & 0.89          & 0.95          & \textbf{0.86} & 0.96          & 0.82          & 0.96          & \textbf{0.85} & 0.95$\pm$0.002                                                     & \textbf{0.86}$\pm$\textbf{0.025} \\
LLaMA              & 0.93          & 0.86          & 0.94          & 0.84          & 0.93          & 0.76          & 0.94          & 0.79          & 0.94$\pm$0.003                                                     & 0.81$\pm$0.040                             \\
MR+AS & \textbf{0.98} & \textbf{0.90} & \textbf{0.99} & 0.85          & \textbf{0.99} & \textbf{0.83} & \textbf{0.99} & 0.84          & \textbf{0.99}$\pm$\textbf{0.002} & \textbf{0.86}$\pm$\textbf{0.021} 
\end{tabular}
}
\vspace{-2mm}
\caption{
Benchmark median CS and ABR metrics.
The last column shows averages over all datasets with standard deviations.
The best results are highlighted in {\bf bold}.
}
\label{baseline_comparison}
\end{table}

\paragraph{Model Benchmark}
Next, we evaluate MLMs (BERT, BART) and LLMs (GPT-4o, LLaMA) on all datasets using median CS and ABR as metrics (distributions are shown in the Appendix Figures~\ref{fig:bench_cs} and \ref{fig:bench_abr}).
Table~\ref{baseline_comparison} shows that MR+AS outperforms both MLMs (BERT, BART) and LLMs (GPT-4o, LLaMA), with top performances in terms of CS in all datasets and top ABR in two of the datasets (SWS and LS14).
Also, when accounting for results variation across datasets, we see that MR+AS significantly outperforms the others in terms of CS, while being comparable with GPT-4o in terms of ABR.
Importantly, for the latter, MR+AS and GPT-4o perform markedly better than other approaches.
%
%

\begin{table}[t]
\centering
\scalebox{0.7}{
\begin{tabular}{c|ccc}
    & CA             & CD             & OMC           \\
    \hline
    BERT-naive & 0.646          & 0.389          & 0.612         \\
    BERT-spsv    & 0.709          & 0.442          & 0.491         \\
    BERT-SWS     & 0.204          & 0.118          & 0.078         \\
    BART-SWS     & 0.225          & 0.139          & 0.121         \\
    Rule Based   & 0.602          & 0.241          & 0.344         \\
    GPT-4o        & 0.761          & 0.563          & 0.661         \\
    LLaMA         & 0.700          & 0.517          & 0.603         \\
    MR+AS         & \textbf{0.961} & \textbf{0.917} & \textbf{0.810}
\end{tabular}
}
\vspace{-2mm}
\caption{Proportion of $p$-values below the significance threshold ($\alpha=0.01$) on the SWS test data.
}
\label{table_benchmark}
\end{table}

\begin{figure}[H] 
    \centering
    \includegraphics[width=\columnwidth]{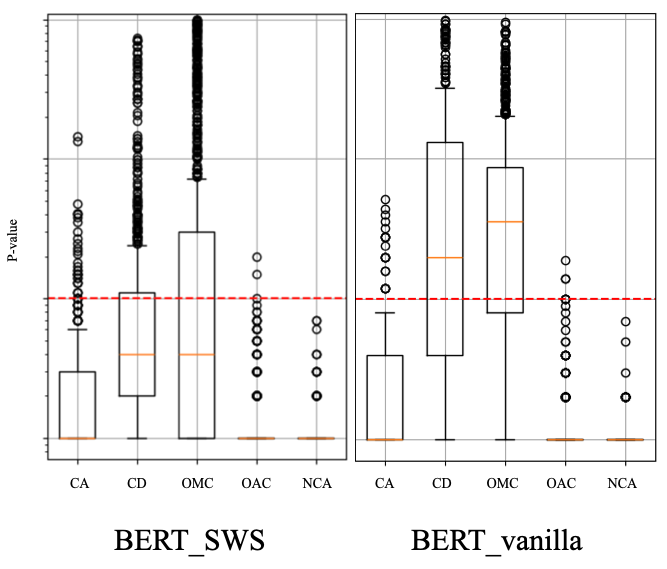} 
    \caption{$p$-value distributions.} 
    \label{fig:benchmark1} 
\end{figure}

\begin{figure}[H] 
    \centering
    \includegraphics[width=\columnwidth]{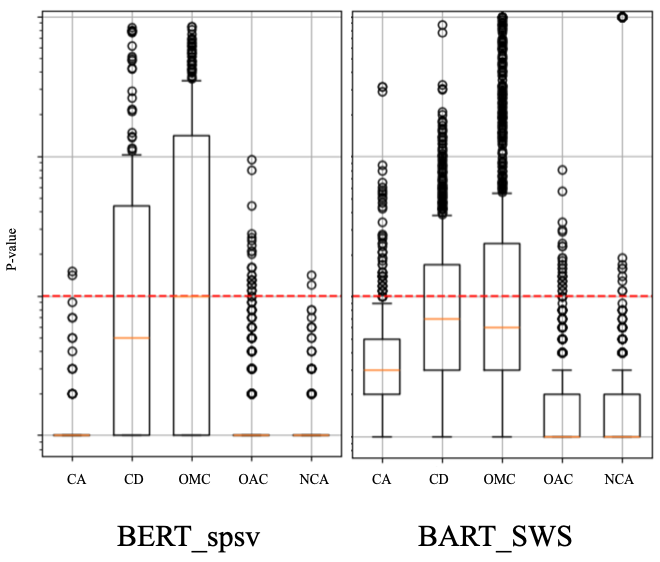} 
    \caption{$p$-value distributions.} 
    \label{fig:benchmark2} 
\end{figure}

\begin{figure}[H] 
    \centering
    \includegraphics[width=\columnwidth]{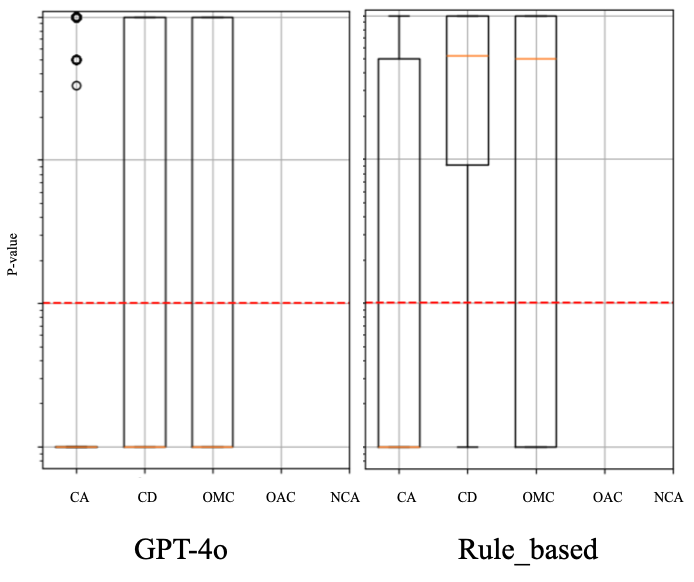} 
    \caption{$p$-value distributions.} 
    \label{fig:benchmark3} 
\end{figure}

\begin{figure}[H] 
    \centering
    \includegraphics[width=\columnwidth]{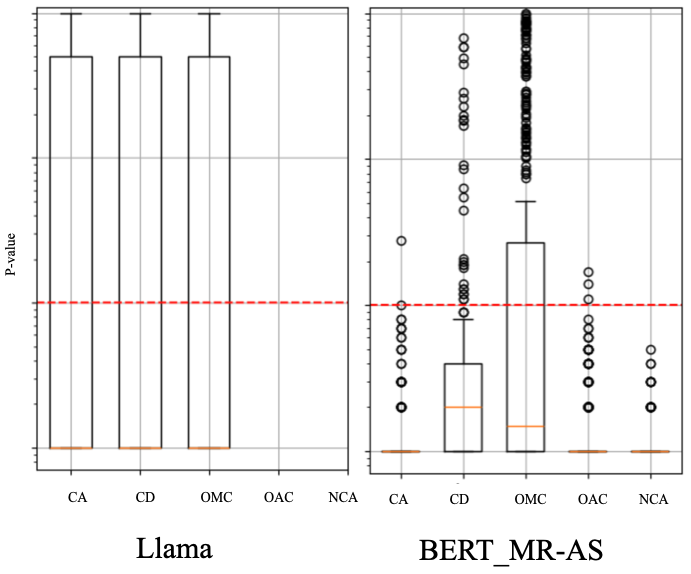} 
    \caption{$p$-value distributions.} 
    \label{fig:benchmark4} 
\end{figure}

\paragraph{Model-based Statistic Benchmark}
After showing the effectiveness of our model in terms of CS and ABR, we now use the model-based statistic introduced in Section~\ref{sc:model-based_stat}.
Table~\ref{table_benchmark} shows the proportion of model predictions that meet the significance threshold ($\alpha=0.01$) for three groups of tokens: CA, CD and OMC.
We do not present results for the other two groups (NCA and OAC) because when using rule- and LLM-based models, we get no candidates for non-substitutions.
Further, for these models we may not always get the same number of substitute candidates per token, thus we use their median (3 candidates) across all tokens in the SWS test set as the number of candidates for all other models when calculating the statistic, {\em i.e.}, $K_s=3$ in \eqref{eq:p-value}.
In the Appendix Table~\ref{tb:pv_all} we show an extended table showing the NCA and OAC groups and $K_s=1000$ for MLM-based models (ours included) and Figures~\ref{fig:benchmark1}-\ref{fig:benchmark4} with $p$-value distributions for all models.
To avoid problems associated with overfitting, thus making the evaluation more reliable, in Appendix Table~\ref{tb:pv_gptscore_GPT2-M} and Table~\ref{tb:pv_gptscore_OPT350M} we extended the evaluation result by using the GPTScore\footnote{Similar to BARTScore, we adapted GPTScore for the paraphrasing task using the following prompt: “Rewrite the following text with the same semantics. \{original sentence\} In other words, \{the modified sentence\}”. We also calculated the Spearman correlation between BARTScore and GPTScore (GPT2-medium and OPT-350M), which yielded results of 0.929 and 0.921, respectively. These values indicate a strong correlation between BARTScore and GPTScore.} \citep{fu2023gptscore} with two backbone models: GPT2 \citep{radford2019language} and OPT \citep{zhang2022opt}. The results show that our MR+AS still outperforms other baseline models.

Ideally, we want proportions in Table~\ref{table_benchmark} to be close to one, indicating that the top candidate is of better quality than the alternatives.
Effectively, results demonstrates that our model (MR+AS) not only produces the largest proportions, but it does so irrespective of the group (CA, CD or OMC).
This suggests that substitutions made by our model are of good quality (using BARTScore as proxy for quality) even if they do not agree with the human annotator.
Moreover, in general, LLMs are better than the other MLM- and rule-based alternatives.

It is worth noting that LLM-based models make considerably more substitutions than our model.
In the Appendix Table~\ref{p-value-proportion-full} we show that GPT-4o makes 24.4\%, 24.8\% and 11.5\% changes in groups CA, CD and OMC, respectively, compared to 5.5\%, 5.2\% and 2.7\% by our MR+AS.
This suggests a trade-off between quality and quantity of substitutions between MR+AS and GPT-4o, respectively.

\begin{table}[t]
\centering
\scalebox{0.7}{
\begin{tabular}{c|cccc|r}
    & SWS            & LS07           & LS14           & XSum           & AVG \\
    \hline
    BERT-naive      & 0.88          & 0.84          & 0.82          & 0.81          & 0.84$\pm $ 0.025                                                      \\
    BERT-spsv         & 0.82           & 0.80          & 0.73          & 0.73          & 0.77$\pm $ 0.040                                                       \\
    BERT-SWS         & 0.85          & 0.82          & 0.78           & 0.79          & 0.81$\pm $ 0.028                                                      \\
    BART-SWS         & 0.82          & 0.81          & 0.74          & 0.77           & 0.79$\pm $ 0.035                                                      \\
    Rule Based        & 0.89           & 0.84          & 0.82          & 0.85          & 0.85$\pm $ 0.025                                                      \\
    GPT-4o             & 0.88          & 0.86          & 0.80          & 0.85          & 0.85$\pm $ 0.028                                                      \\
    LLaMA              & 0.86           & 0.83          & 0.77          & 0.79          & 0.81$\pm $ 0.035                                                      \\
    MR+AS & \textbf{0.91} & \textbf{0.87} & \textbf{0.85} & \textbf{0.85} & \textbf{0.87} $\pm $ \textbf{0.024}
\end{tabular}}
\vspace{-2mm}
\caption{Median BARTScore ratios for top-2 candidates.
The last column shows averages over all datasets with SDs.
The best results are highlighted in {\bf bold}.
}
\label{figure_top2}
\end{table}

\paragraph{Top-2 Candidate Performance}
Provided that the substitution rates by our model are low relative to others as discussed above, we now examine the quality of the top-2 substitute.
Note that in MLM-based models a change is produced only when the top substitute candidate is different than the original token and that for LLM-based models the top-2 candidate is only available is the model considers a substitution has to be made and more than one candidates are provided.
To quantify the quality of the top-2 candidates, we randomly selected five tokens from each sentence where the top-2 candidate is available and calculate the BARTScore ratio between the sentence modified with the top-candidate and that of the original (unchanged) sentence.
We then averaged these ratios provide a summary of the quality of top-2 candidates.
Results in Table~\ref{figure_top2} indicate that MR+AS achieved the best top-2 candidate quality across all four datasets followed (on average) by GPT-4o and the rule-based model.
%

\paragraph{Encouraging ranking in LLMs}
It is well known that prompting LLMs to provide ordered responses is challenging \citep{lu2023error}.
We tried prompts with and without a specific request for ordered substitute candidates (see Appendix~\ref{sc:llm_prompts} for details).
We found that both GPT and LLaMA produced slightly better median CS values without the prompt specifying ranked candidates (0.954 and 0.943, respectively) relative to the alternative (0.953 and 0.940, respectively).
Importantly, these are significantly lower than the median CS value for MR+AS (0.983).
CS distributions for all models can be found in the Appendix Figure~\ref{fig:llama/gpt_order}.
We note that it may be possible to optimize the prompt for better ranked results, however, this is beyond the scope of our work.

\paragraph{Human Study}
We recruited five annotators to carry out the study.
First, we randomly sampled 25 cases from the SWS test data and provided the top two candidate replacements proposed by our model. 
We randomly flipped the order of these two candidates before presenting them to the annotators.
Annotators were asked to indicate their preference for the order and whether they preferred to replace the target word with either of the two candidates\footnote{Example of questionnaire 1 can be accessed at \href{https://docs.google.com/spreadsheets/d/1G__gBcVt1bDd5UNVrEgdqmPMItpvOJe3j8mKAdkJrKg/edit?usp=sharing}{https://tinyurl.com/SWSQuestionnaire1}.}.

To assess inter-annotator agreement, we calculated the (unweighted) Cohen’s kappa coefficient for both order agreement and replacement agreement.
The average kappa values for the order agreement and the replacement agreement were 0.40$\pm0.21$ (standard deviation) and 0.13$\pm$0.21, respectively.
These results align with our assumption that asking annotators to suggest replacements for a specific word in a given sentence is a subjective task.
In Appendix Table~\ref{kappa}, we present the complete results. 
Furthermore, we observed that in 83\% of the cases, the annotators did not choose to replace the target word.

To mitigate the potential bias caused by providing the target word in advance, we designed a second questionnaire consisting of 50 cases.
Twenty-five cases were sampled where our model and BARTScore agreed, and twenty-five cases where they disagreed.
In this setting, the target word was hidden and only the top two candidate replacements (with randomly flipped order) were provided. 
Annotators were then asked to choose their preference from three options\footnote{Example of questionnaire 2 can be accessed at \href{https://docs.google.com/spreadsheets/d/1SmkAxzvflaq6Qhm4DB37LQu24pS1B4ATBGwzMSXyqkk/edit?usp=sharing}{https://tinyurl.com/SWSQuestionnaire2}.}:
(0) the top-1 and top-2 candidates are equally good.
(1) the top-1 candidate is better.
(2) the top-2 candidate is better.

Subsequently, we made a comparison between the model predictions and the responses of the annotators.
The results showed that in 38.8\% of the the instances, annotators considered both candidates equally good.
In the remaining cases, when the model and BartScore agreed, 79\% of annotators favored the same choice.
In cases where the model and BartScore disagreed, annotators sided with BartScore 67\% of the time and with the model 33\% of the time. In the SWS test data, the model and BartScore agreed on the ranking of the top two candidates in 67\% of instances.
In Appendix Table \ref{human-study}, we present the complete results.

\section{Discussion}
This paper addressed the problem of training and evaluating SWS models without requiring human annotations.
Specifically, we leveraged a model-based score (BARTScore) to define evaluation metrics and to serve as a label-proxy for our loss function.
Extensive experimental results demonstrate the effectiveness of our approach relative to both MLM- and LLM-based models.
%


\section*{Limitations}
The proposed method leverages BARTScore as a sentence scoring mechanism, essentially acting as a proxy for human annotators in guiding model optimization during preference-aware learning.
However, we cannot eliminate the possibility that BARTScore may not always accurately quantify the quality of token substitutions.
While BARTScore has shown effectiveness in our experiments, it could be readily replaced by alternative scoring metrics deemed more reliable or better suited to other word substitution contexts.
We leave this exploration as interesting future work.

\section*{Ethical Considerations}
In the process of developing and deploying smart word suggestion systems it is crucial to consider the broader ethical implications of relying on automated tools for language optimization.
These systems may inadvertently reinforce biases present in training data, leading to unintended or inappropriate word suggestions.
Moreover, the replacement of human judgment with automated scoring models like BARTScore can risk overlooking or misrepresenting nuances in language, particularly in sensitive contexts.
Therefore, we recommend ongoing human oversight and continuous evaluation of the model's output to ensure fairness, inclusivity, and alignment with ethical standards.




\clearpage

\appendix

\section{DPO Derivation Details}\label{sc:dpo_det}
\paragraph{DPO*}
Direct Preference Optimization (DPO) \citep{rafailov2024direct} is an efficient technique for aligning large language models (LLMs) with human feedback, which gained popularity due to its simplicity \citep{miao2024inform}.
For instance, it has demonstrated to be effective in chat benchmarks \citep{tunstall2023zephyr, zheng2023judging}.
In our case, the model-based score serving as proxy for human feedback
DPO (under the Plackett-Luce Model) is written as
\begin{align}\label{eq:dpo_}
\mathcal{L}_{\rm DPO} = -\mathbb{E} 
\left[ \log \textstyle{\prod}_{k=1}^{K} 
\frac{\exp ( \delta  r_k )}{ \sum_{j=k}^{K} \exp ( \delta r_j )} \right] \,, 
\end{align}
where $r_k=\log p_\theta(\tilde{X}_k) - \log p_{\hat{\theta}}(\tilde{X}_k)$, the expectation is over $\{\tilde{X}_1, \ldots, \tilde{X}_K, X\}$, and 
$\theta$ and $\hat{\theta}$ denote the parameters of the model being trained and that used for reference, respectively.
Accordingly, only $\theta$ are updated while learning while $\hat{\theta}$ are kept fixed.

Since the magnitude of the logits drops significantly as $k\to K$, we found that the sum in the denominator of \eqref{eq:dpo_} weakens the loss.
Therefore, we removed the sum and instead compared the $k$-th and the $(k+1)$-th substitution, rather than comparing it with all $K$ values.
The Derivation of DPO* is shown below
\begin{align}
\mathcal{L}_{\text{DPO*}} &= - \log \textstyle{\prod}_{k=1}^{K-1}  \frac{\exp \left( \delta  r_k  \right)}{\exp \left( \delta  r_{k+1}  \right)} \label{eq:dpo*_derivation} \\
&= -\textstyle{\sum}_{k=1}^{K-1}\log \frac{\exp \left( \delta  r_k  \right)}{\exp \left( \delta  r_{k+1} \right)} \notag \\
&= -\textstyle{\sum}_{k=1}^{K-1} \left( \delta  r_k  - \delta  r_{k+1}  \right) \notag \\
&= -\textstyle{\sum}_{k=1}^{K-1} \delta\left( \log \frac{p_\theta(\tilde{X}_k)}{p_{\hat{\theta}}(\tilde{X}_k)} - \log \frac{p_\theta(\tilde{X}_{k+1})}{p_{\hat{\theta}}(\tilde{X}_{k+1})} \right) \,. \notag
\end{align}
Further, we approximate $\log p_\theta(\tilde{X}_k)$ with its logit $s_k$, and let $\delta=1$ for simplicity.
Then \eqref{eq:dpo*_derivation} simplifies (for a single token in sentence $X$) to
\begin{align*}
    \mathcal{L}_{\text{DPO*}} &= -\textstyle{\sum}_{k=1}^{K-1} \left( s_k - \hat{s}_k - s_{k+1} + \hat{s}_{k+1} \right) \,, 
\end{align*}
where $s_k$ and $\hat{s}_k$ denote the logit of the $k$-th substitution from the model being trained and the reference model, respectively.

\paragraph{$\sigma$DPO*}
We also extend DPO (under the Bradley-Terry model) \citep{rafailov2024direct} to multiple substitute candidates, where each candidate is compared with the next in the ordered list of candidates.
We write for a single token in sentence $X$
\begin{dmath}
\mathcal{L}_{\sigma\mathrm{DPO*}}=- \textstyle{\sum}_{k=1}^{K-1}\log \sigma\left(\delta  r_k-\delta  r_{k+1} \right) \,.
\label{eq:sigma_dpo*_derivation}
\end{dmath}
We approximate $\log p_\theta(\tilde{X}_k)$ with its logit $s_k$, and let $\delta=1$ for simplicity.
Then in \eqref{eq:sigma_dpo*_derivation} simplifies (for a single token in sentence $X$) to
\begin{align*}
& \mathcal{L}_{\text{$\sigma$DPO*}} = \\
& \hspace{8mm} -\textstyle{\sum}_{k=1}^{K-1} \log\sigma \left( s_k - \hat{s}_k - s_{k+1} + \hat{s}_{k+1} \right) \,,
\end{align*}
from which we see that the only difference between \eqref{eq:dpo*-sigma} and \eqref{eq:dpo*} (in the main paper) is that the comparison of logit values in the former is scaled with the log-logistic function.
See Appendix for a derivation of both losses.

\section{LLM Prompts}\label{sc:llm_prompts}
Prompt \underline{without} order:

\textit{In the following sentence, please give some suggestions to improve word usage. Please give the results with the JSON format of {“original word”: [“suggestion 1”, “suggestion 2”]}. The 'original word' should include all words that can be improved in the sentence, directly extracted from the sentence itself. [s]}

\noindent Prompt \underline{with} order:

\textit{In the following sentence, please give some suggestions to improve word usage. Please give the results with the JSON format of {“original word”: [“suggestion 1”, “suggestion 2”]}. The 'original word' should include all words that can be improved in the sentence, directly extracted from the sentence itself, and \textbf{the suggestions should be ranked in order of the degree of improvement, from the most effective to the least}. [s]}

where [s] is the sentence. 

At times, the model may fail to provide the correct JSON format, making it difficult to extract the intended answer.
In such cases, it is often necessary to query the model multiple times to obtain a valid JSON output.

\section{Model Hyperparameters}\label{sc:hyperparams}
Both of BERT\_SWS reproduction and model optimization are: $\text{epochs}=5$, $\text{batch size} = 64$, $\text{learning rate} = 0.0007$, $\text{max norm} = 1e-5$ for clipped gradient norm, $\text{dropout rate} = 0.1$, $\lambda = 0.5$, and $\gamma = 1$.

\begin{table}[h]
\centering
\scalebox{0.73}{
\begin{tabular}{lllll}
\hline
Data & Task                 & Training  & Validation & Test   \\ \hline
SWS  & SWS                  & 3,909,650 & 200        & 800    \\
LS07 & Lexical Substitution & -         & 299        & 1710   \\
LS14 & Lexical Substitution & -         & 892        & 1569   \\
XSum & Summarization        & 204045    & 11332      & 11,334 \\ \hline
\end{tabular}}
\caption{Datasets summary.
}
\label{tb:data_summ}
\end{table}

%

\begin{table*}[t]
\centering
\scalebox{0.55}{
\begin{tabular}{lp{23cm}}
\multirow{35}{*}{CD} & \begin{tabular}[c]{@{}l@{}}While being saddened by my friend's suffering, I also gradually became aware of the  \textcolor{red}{critical} role of law in our daily life.\\ Annotations: Human=\{"integral", "important"\}, Model=\{"crucial"\}\end{tabular} \\
\cline{2-2} 
& \begin{tabular}[c]{@{}l@{}}The residential theory \textcolor{red}{argues} that there are houses in the Chaco structures, and the structure is big enough for hundreds of people. \\ Annotations: Human=\{"discussed", "states",   "reasons", "discusses"\}, Model=\{"claims"\}\end{tabular} \\
\cline{2-2}
& \begin{tabular}[c]{@{}l@{}}As I planned   well for the whole day, I would not make mistakes such as doing things relating to a single subject in a \textcolor{red}{complete} morning or afternoon, \\which would make my head dizzy and I  could not focus all my attention on my work.\\      Annotations: Human=\{"single", "whole"\}, Model= \{"full"\}\end{tabular} \\  
\cline{2-2} 
& \begin{tabular}[c]{@{}l@{}}By contrast, the professor refutes this idea   and \textcolor{red}{insists} that it will not   be that profitable when the cost is taken into consideration.\\      Annotations: Human=\{"urges", "claims"\} Model=  \{"argues"\}\end{tabular}  \\
\cline{2-2} 
& \begin{tabular}[c]{@{}l@{}}For example, in traditional China, young people   needed to turn down their knees to \textcolor{red}{show} their respect to old people like their parents or their   teachers.\\      Annotations: Human=\{"convey", "display",   "manifest"\} Model= \{"demonstrate"\}\end{tabular}   \\
\cline{2-2} 
& \begin{tabular}[c]{@{}l@{}}As a matter of fact, young people didn't need   to obey too many strict rules and could have a \textcolor{red}{relatively} relaxing life.\\      Annotations: Human=\{"generally", "decidedly",   "approximately"\} Model= \{"fairly"\}\end{tabular}     \\ 
\cline{2-2}
& \begin{tabular}[c]{@{}l@{}}To be more specific, most young men tend to be   couch potatoes during weekends, leading to not \textcolor{red}{concerning}  about social issues.\\      Annotations: Human=\{"updating",    "concerns", "regarding",  "caring"\} Model=  \{"worrying"\}\end{tabular}  \\
\cline{2-2} 
& \begin{tabular}[c]{@{}l@{}}As a result, to make sure that I fulfil the \textcolor{red}{goal} I made every day, I would   prevent myself from doing  unnecessary things like playing for a long time in \\  the playground as some of my classmates would do.\\      Annotations: Human=\{"aim"\} Model= \{"objective"\}\end{tabular}  \\   \cline{2-2} 
& \begin{tabular}[c]{@{}l@{}}When enjoying this piece, people can directly   follow the progress of the piece to feel the picture constructed between the \textcolor{red}{various} instruments \\without   exploring its creative background.\\      Annotations: Human=\{"diverse"\} Model= \{"different"\}\end{tabular}    \\ \cline{2-2}
& \begin{tabular}[c]{@{}l@{}}No one \textcolor{red}{likes} to deal with people who can't adopt other people's advice and   insist that only his or her idea  is correct.\\      Annotations: Human=\{"prefers", "enjoys"\} Model=  \{"wants"\}\end{tabular}  \\ \cline{2-2}
& \begin{tabular}[c]{@{}l@{}}Globalists view the coronavirus as a global   threat that shows the \textcolor{red}{common}   plight of humanity and the need to work together.\\      Annotations: Human=\{"reoccurring", "everyday"\} Model=  \{"shared"\}\end{tabular}      \\ \cline{2-2} 
& \begin{tabular}[c]{@{}l@{}}Second, the reading \textcolor{red}{suggests} that the mining industry on asteroids would be highly   profitable due to the numerous valuable elements and precious\\ metals buried   under the asteroids.\\      Annotations: Human=\{"implies", "proposes"\} Model=  \{"indicates"\}\end{tabular}    \\ \hline
\multirow{35}{*}{OMC} & \begin{tabular}[c]{@{}l@{}}Therefore, the power to \textcolor{red}{resolve} disputes and maintain social order is the psychological  restraint of the actor and the education of the elder to the \\young rather than the litigation.\\ Annotations: Human=\{"resolve"\}, Model=\{"settle"\}\end{tabular} \\
\cline{2-2} 
& \begin{tabular}[c]{@{}l@{}}According to recent statistics, there is ample coverage demonstrating that humans who inhabit the hub \textcolor{red}{continually} will undergo severe conditions due\\ to the dirty air and stressful surroundings.\\ Annotations: Human =\{"continually"\}, Model=\{"constantly"\}\end{tabular} \\ \cline{2-2}
& \begin{tabular}[c]{@{}l@{}}We are capable of making crops which can \textcolor{red}{resist} many kinds of unideal   factors, such as colds, bugs, or even drought.\\      Annotations: Human=\{resist\} Model= \{"withstand"\}\end{tabular}   \\ \cline{2-2}
& \begin{tabular}[c]{@{}l@{}}With the proliferation of schools and private   firms nowadays, it is \textcolor{red}{sometimes}   argued that a company or campus should have strict rules that control\\ the   type of clothing that people wear at work and at school.\\      Annotations: Human=\{sometimes\} Model= \{"occasionally"\}\end{tabular}   \\ \cline{2-2} 
& \begin{tabular}[c]{@{}l@{}}The number of students who have grades over 90   in international exams this year has surprisingly soared roughly about 10   percent, a figure that has \\ \textcolor{red}{nearly} doubled as against that of last year.\\      Annotations: Human=\{nearly\} Model= \{"almost"\}\end{tabular}    \\ \cline{2-2}
& \begin{tabular}[c]{@{}l@{}}As a \textcolor{red}{result}, to make sure that I fulfil the goal I made every day, I would   prevent myself from doing unnecessary things like playing for a long time \\in   the playground as some of my classmates would do.\\      Annotations: Human=\{result\} Model= \{"consequence"\}\end{tabular}     \\ \cline{2-2}
& \begin{tabular}[c]{@{}l@{}}Even though the issue is becoming increasingly   critical, many believe it is not \textcolor{red}{essential} for the young to learn how to plan and organize.\\      Annotations: Human=\{essential\} Model= \{"necessary"\}\end{tabular}   \\ \cline{2-2}
& \begin{tabular}[c]{@{}l@{}}When enjoying this piece, people can directly   follow the progress of the piece to feel the picture constructed between the   various instruments \\without \textcolor{red}{exploring} its creative background.\\      Annotations: Human=\{exploring\} Model= \{"examining"\}\end{tabular}     \\ \cline{2-2} 
& \begin{tabular}[c]{@{}l@{}}There is no doubt that letting students decide   how many and how \textcolor{red}{often} they could do study assignments is more flexible.\\      Annotations: Human=\{often\} Model= \{"frequently"\}\end{tabular}  \\ \cline{2-2} 
& \begin{tabular}[c]{@{}l@{}}This is because the Parliament has the   intention to limit personal liberty identified as fundamental rights by the   unequivocal language without \\ \textcolor{red}{allowing} the diagnosed person to leave the residence or places except   for certain circumstances (Momcilovic; Aboriginal Justice; s4).\\      Annotations: Human=\{allowing\} Model= \{"permitting"\}\end{tabular} \\ \cline{2-2}
& \begin{tabular}[c]{@{}l@{}}Those who bolster the idea might indicate this   operation either is interesting to allow them to engage in a relaxed   atmosphere or spend less time\\ \textcolor{red}{focusing} on solid books.\\      Annotations: Human=\{focusing\} Model= \{"concentrating"\}\end{tabular} \\ \cline{2-2}
& \begin{tabular}[c]{@{}l@{}}What   is more, if young students keep making use of video games in some lessons,   their caliber of watching objects will \textcolor{red}{decrease}, granted that \\they avail themselves of exhibiting sight too   much.\\      Annotations: Human=\{decrease\} Model= \{"decline"\}\end{tabular} \\
\end{tabular}
}
\caption{Original sentences with target token substitution in \textcolor{red}{red}. CD: change disagreement and OMC: only model changed.
}
\label{tb:cases}
\end{table*}

\begin{figure}[H]
    \centering
    \includegraphics[width=\columnwidth]{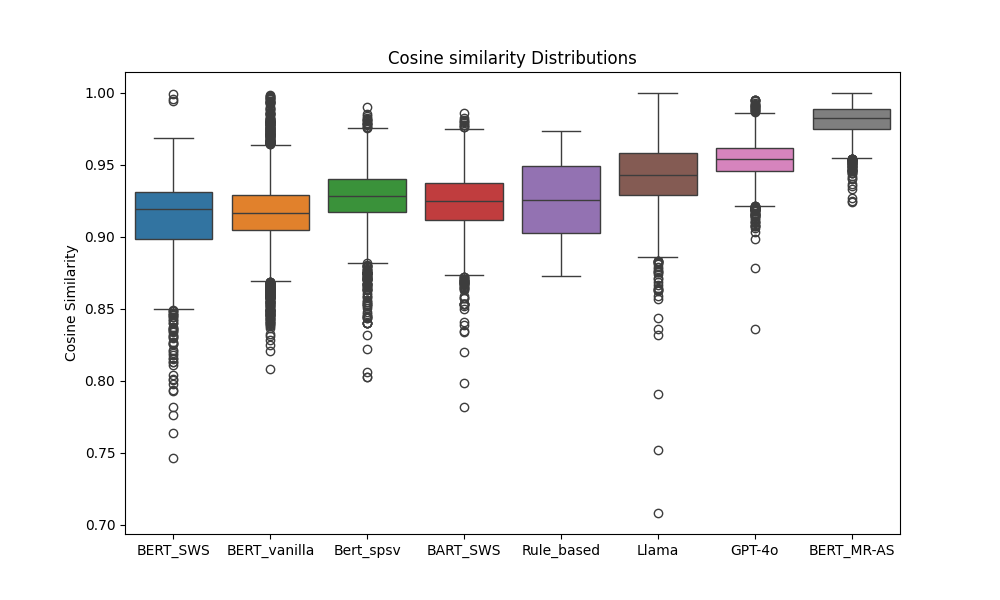}
    \caption{CS distribution results for SWS test data.
    }
    \label{fig:bench_cs}
\end{figure}

\begin{figure}[H]
    \centering
    \includegraphics[width=\columnwidth]{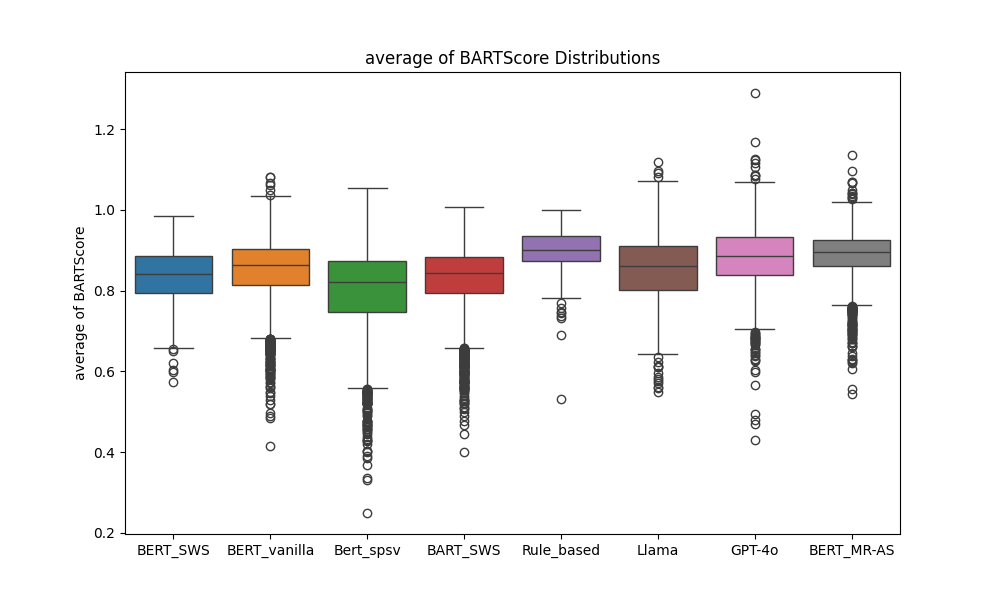}
    \caption{ABR distribution results for SWS test data.
    }
    \label{fig:bench_abr}
\end{figure}


\begin{table}[H]
\centering
\scalebox{0.8}{
\begin{tabular}{c|c|c|c|c|c}
                  & OAC                           & CA                              & CD                          & OMC                           & NCA                    \\ \hline
BERT-naive     & \textbf{0.999} & 0.888                           & 0.339                         & 0.274                           & \textbf{1} \\
BERT-spsv        & 0.994                           & 0.981                           & 0.583                         & 0.486                           & 0.999                       \\
BERT-SWS         & \textbf{0.999} & 0.942                           & 0.731                         & 0.628                           & \textbf{1} \\
BART-SWS         & 0.992                           & 0.905                           & 0.604                         & 0.588                           & 0.969                       \\
Rule Based       & -                               & 0.602                           & 0.241                         & 0.344                           & -                           \\
GPT-4o            & -                               & 0.761                           & 0.563                         & 0.661                           & -                           \\
LLaMA             & -                               & 0.700                           & 0.517                         & 0.603                           & -                           \\
MR+AS & \textbf{0.999} & \textbf{0.994} & \textbf{0.9} & \textbf{0.705} & \textbf{1}
\end{tabular}}
\caption{Proportion of $p$-values below the significance threshold ($\alpha=0.01$) for $K_s=1000$ using BARTScore.
}
\label{tb:pv_all}
\end{table}

\begin{table}[H]
\centering
\scalebox{0.8}{
\begin{tabular}{c|c|c|c|c|c}
           & OAC        & CA             & CD             & OMC            & NCA        \\ \hline
BERT-naive & \textbf{1} & 0.888          & 0.425          & 0.47           & \textbf{1} \\
BERT-spsv  & \textbf{1}          & 0.891          & 0.474          & 0.438          & \textbf{1}          \\
BERT-SWS   & \textbf{1} & 0.948          & 0.717          & 0.621          & \textbf{1} \\
BART-SWS   & \textbf{1}          & 0.869          & 0.6            & 0.595          & \textbf{1}          \\
Rule Based & -          & 0.552          & 0.248          & 0.309          & -          \\
GPT-4o     & -          & 0.734          & 0.557          & 0.655          & -          \\
LLaMA      & -          & 0.654          & 0.518          & 0.624          & -          \\
MR+AS      & \textbf{1} & \textbf{0.963} & \textbf{0.809} & \textbf{0.719} & \textbf{1}
\end{tabular}}
\caption{Proportion of $p$-values below the significance threshold ($\alpha=0.01$) for $K_s=1000$ using GPTScore with GPT2-medium (355M).
}
\label{tb:pv_gptscore_GPT2-M}
\end{table}

\begin{table}[H]
\centering
\scalebox{0.8}{
\begin{tabular}{c|c|c|c|c|c}
           & OAC        & CA             & CD             & OMC            & NCA        \\ \hline
BERT-naive & \textbf{1} & 0.893          & 0.439          & 0.351          & \textbf{1} \\
BERT-spsv  & \textbf{1}          & 0.927          & 0.506          & 0.498          & \textbf{1}          \\
BERT-SWS   & \textbf{1} & 0.975          & 0.749          & 0.655          & \textbf{1} \\
BART-SWS   & \textbf{1}          & 0.897          & 0.618          & 0.625          & \textbf{1}          \\
Rule Based & -          & 0.531          & 0.247          & 0.332          & -          \\
GPT-4o     & -          & 0.745          & 0.554          & 0.673          & -          \\
LLaMA      & -          & 0.694          & 0.531          & 0.625          & -          \\
MR+AS      & \textbf{1} & \textbf{0.981} & \textbf{0.759} & \textbf{0.657} & \textbf{1}
\end{tabular}}
\caption{Proportion of $p$-values below the significance threshold ($\alpha=0.01$) for $K_s=1000$ using GPTScore with OPT-350M (350M).
}
\label{tb:pv_gptscore_OPT350M}
\end{table}

\begin{table}[H]
\centering
\scalebox{0.7}{
\begin{tabular}{c|c|c|c|c|c|c}
              & OAC            & CA    & CD  & OMC   & NCA           & Ratio         \\ \hline
BERT-naive & 0.914          & 0.032 & 0.054 & 0.052 & 0.948          & 4.09          \\
BERT-spsv    & 0.952 & 0.02  & 0.028 & 0.017 & 0.983 & 2.82          \\
BERT-SWS     & 0.738          & 0.113 & 0.149 & 0.064 & 0.936          & 1.65          \\
BART-SWS     & 0.736          & 0.113 & 0.151 & 0.068 & 0.932          & 3.88          \\
Rule Based   & 0.63           & 0.11  & 0.26  & 0.101 & 0.899          & 3.66          \\
GPT-4o        & 0.508          & 0.244 & 0.248 & 0.115 & 0.885          & 4.28 \\
LLaMA         & 0.722          & 0.121 & 0.157 & 0.065 & 0.935          & 4.28 \\
MR+AS         & 0.893          & 0.055 & 0.052 & 0.027 & 0.973          & 3.96         
\end{tabular}}
\caption{The value indicates the proportion of tokens involved. We divided the data into three groups (OAC, CA, CD) for changes made by human annotators, and two groups (OMC, NCA) for tokens the human annotator did not change.
The ratio in the last column is between tokens both the model and the annotator change (CA and CD) and tokens only the model changed (NCA).
}
\label{p-value-proportion-full}
\end{table}

\begin{figure}[H]
    \centering
    \includegraphics[width=\columnwidth]{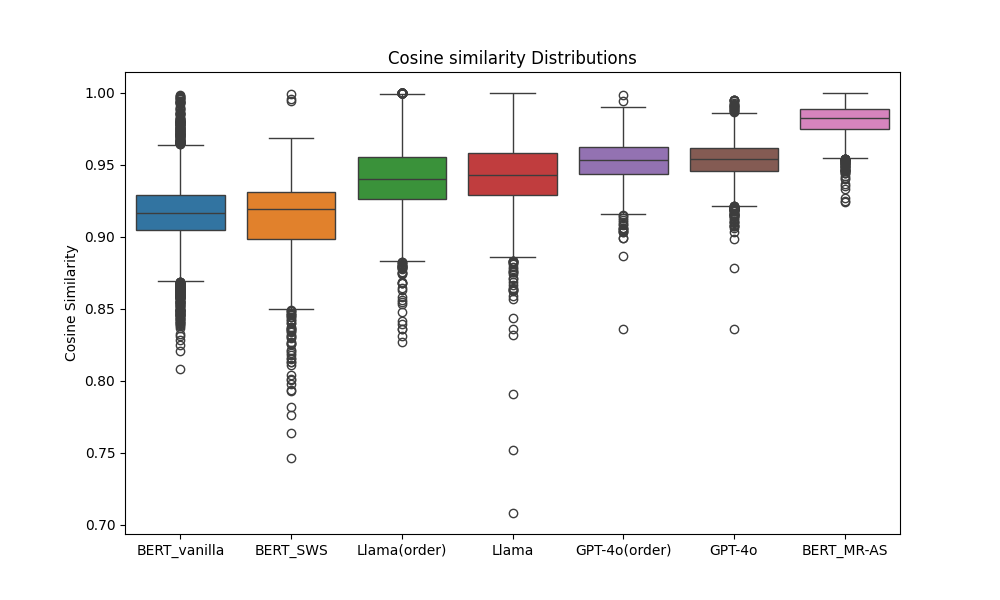}
    \caption{Results for LLaMA and GPT-4o with and without a prompt encouraging ranking.
    }
    \label{fig:llama/gpt_order}
\end{figure}

\begin{table}[H]
\centering
\scalebox{0.57}{
\begin{tabular}{l|rrrrr}
\multicolumn{6}{c}{Order Agreement}                                                                                                               \\
      & \multicolumn{1}{l}{Annotator1} & \multicolumn{1}{l}{Annotator2} & \multicolumn{1}{l}{Annotator3} & \multicolumn{1}{l}{Annotator4} & \multicolumn{1}{l}{Annotator5} \\
      \hline
Annotator1 & -                         & 0.28                      & 0.51                      & 0.75                      & 0.44                      \\
Annotator2 & -                         & -                         & 0.12                      & 0.19                      & 0.2                       \\
Annotator3 & -                         & -                         & -                         & 0.41                      & 0.44                      \\
Annotator4 & -                         & -                         & -                         & -                         & 0.68                      \\
Annotator5 & -                         & -                         & -                         & -                         & -                         \\
Average$\pm$SD   & \multicolumn{1}{l}{}      & \multicolumn{1}{l}{}      & \multicolumn{1}{l}{}      & \multicolumn{1}{l}{}      & 0.40$\pm$0.21                     \\
\multicolumn{6}{c}{Replacement Agreement}                                                                                                         \\
      & \multicolumn{1}{l}{Annotator1} & \multicolumn{1}{l}{Annotator2} & \multicolumn{1}{l}{Annotator3} & \multicolumn{1}{l}{Annotator4} & \multicolumn{1}{l}{Annotator5} \\
      \hline
Annotator1 & -                         & -0.03                     & -0.2                      & 0.28                      & 0.43                      \\
Annotator2 & -                         & -                         & 0.11                      & 0.17                      & 0.4                       \\
Annotator3 & -                         & -                         & -                         & -0.16                     & 0.11                      \\
Annotator4 & -                         & -                         & -                         & -                         & 0.17                      \\
Annotator5 & -                         & -                         & -                         & -                         & -                         \\
Average$\pm$SD   & \multicolumn{1}{l}{}      & \multicolumn{1}{l}{}      & \multicolumn{1}{l}{}      & \multicolumn{1}{l}{}      & 0.13$\pm$0.21                    
\end{tabular}}
\caption{Inter-annotator agreement using (unweighted) Cohen’s kappa coefficient. SD is the standard deviation.
}
\label{kappa}
\end{table}

\begin{table}[H]
\centering
\scalebox{0.66}{
\begin{tabular}{l|rrr}
\multicolumn{4}{c}{Model and BARTScore Agreement (25 cases)}                  \\
           & Select same          & Select different & Select 0    \\
           \hline
Annotator1 & 7                    & 1                & 17          \\
Annotator2 & 12                   & 5                & 8           \\
Annotator3 & 14                   & 9                & 2           \\
Annotator4 & 18                   & 5                & 2           \\
Annotator5 & 8                    & 0                & 17          \\
Average$\pm$SD        & 11.8$\pm$4.5                 & 4$\pm$3.6                & 9.2$\pm$7.5         \\
\multicolumn{4}{c}{Model and BARTScore Disagreement (25 cases)}               \\
           & Agree with BARTScore & Agree with Model & Select 0    \\
           \hline
Annotator1 & 6                    & 3                & 16          \\
Annotator2 & 9                    & 7                & 9           \\
Annotator3 & 14                   & 7                & 3           \\
Annotator4 & 14                   & 8                & 3           \\
Annotator5 & 4                    & 1                & 20          \\
Average$\pm$SD        & 9.4$\pm$4.6                  & 5.2$\pm$3.0              & 10.2$\pm$7.7        \\
\end{tabular}}
\caption{Human study results. The Model is our (MR+AS) model. SD is the standard deviation. Based on the last column, we calculated the proportion of cases in which humans selected 0 by dividing the last column’s value by 25 and computing the average, resulting in 38.8\%. For the remaining cases (where the selection is either 1 or 2), we analyzed agreement between the annotator, the model, and BARTScore. In the top panel, the proportion of cases where the annotator agree with model and the BARTScore is calculated as the value in the first column divided by the sum of the first and second columns, resulting in an average of 79\%. In the bottom panel, we computed two proportions: the agreement between the annotator and BARTScore, calculated as the value in the first column divided by the sum of the first and second columns, and the agreement between the annotator and the model, calculated as the value in the second column divided by the sum of the first and second columns. The resulting agreement proportions are 67\% for annotator-BARTScore agreement and 33\% for annotator-model agreement.
}
\label{human-study}
\end{table}

\end{document}